%% file: latex/00_main.tex
\pdfoutput=1

\documentclass[11pt]{article}

\usepackage[final]{acl}

\usepackage{times}
\usepackage{latexsym}

\usepackage[T1]{fontenc}

\usepackage[utf8]{inputenc}

\usepackage{microtype}

\usepackage{inconsolata}

\usepackage{graphicx}
\usepackage{amsmath}
\usepackage{amssymb}
\usepackage{booktabs}
\usepackage{pdfpages}
\usepackage{subcaption}
\usepackage{tcolorbox}
\usepackage{geometry}
\usepackage[table]{xcolor}
\usepackage{makecell}
\usepackage{multirow}

\tcbuselibrary{listingsutf8}

\newcommand{\rp}[1]{\textcolor{blue}{RP: #1}}
\newcommand{\mee}[1]{\textcolor{red}{#1}}

\title{CAPO: Confidence Aware Preference Optimization Learning for Multilingual Preferences}

\author{Rhitabrat Pokharel, Yufei Tao, Ameeta Agrawal\\
Department of Computer Science\\
Portland State University, USA \\
\texttt{\{pokharel, yutao, ameeta\}@pdx.edu} \\}

\begin{document}
\maketitle
\begin{abstract}
Preference optimization is a critical post-training technique used to align large language models (LLMs) with human preferences, typically by fine-tuning on ranked response pairs. While methods like Direct Preference Optimization (DPO) have proven effective in English, they often fail to generalize robustly to multilingual settings. We propose a simple yet effective alternative, Confidence-Aware Preference Optimization (CAPO), which replaces DPO’s fixed treatment of preference pairs with a dynamic loss scaling mechanism based on a relative reward. By modulating the learning signal according to the confidence in each preference pair, CAPO enhances robustness to noisy or low-margin  comparisons, typically encountered in multilingual text. Empirically, CAPO outperforms existing preference optimization baselines by at least 16\% in reward accuracy, and improves alignment by widening the gap between preferred and dispreferred responses across languages.

\end{abstract}

\input{latex/01_intro}

\input{latex/02_method}
\input{latex/03_results}
\input{latex/04_related_work}

\input{latex/05_conlcusion}

\section*{Limitations}
We conduct our experiments using LoRA-based fine-tuning with a small training set for each language. While full fine-tuning may yield different insights, our results demonstrate that even with efficient, lightweight training, CAPO achieves meaningful improvements. Additionally, there remain many languages that could not be studied in this work.

\section*{Ethical Consideration}
While our study focuses on a small set of languages, other languages especially low-resource languages remain underexplored and could benefit significantly from improved alignment techniques.

\section*{Acknowledgement}
We thank Dr. Suresh Singh, PortNLP Lab members, and the anonymous reviewers for their constructive feedback.

\bibliography{latex/00_main}

\appendix
\input{latex/06_appendix}

\end{document}

%% file: latex/01_intro.tex
\section{Introduction}


Preference optimization (PO) is a widely adopted post-training technique used to enhance the performance of large language models (LLMs) by aligning their outputs with human preferences. This is typically achieved by fine-tuning models using ranked responses or preference-based signals. While effective, most existing work in this area has been heavily centered on English \cite{rafailov2023direct, meng2024simpo, kto2024, guo-etal-2024-controllable}, with only recent efforts beginning to explore its application in languages other than English.

\begin{figure}[!t]
    \centering
    \begin{subfigure}[b]{0.45\textwidth}
        \centering
        \includegraphics[width=\textwidth]{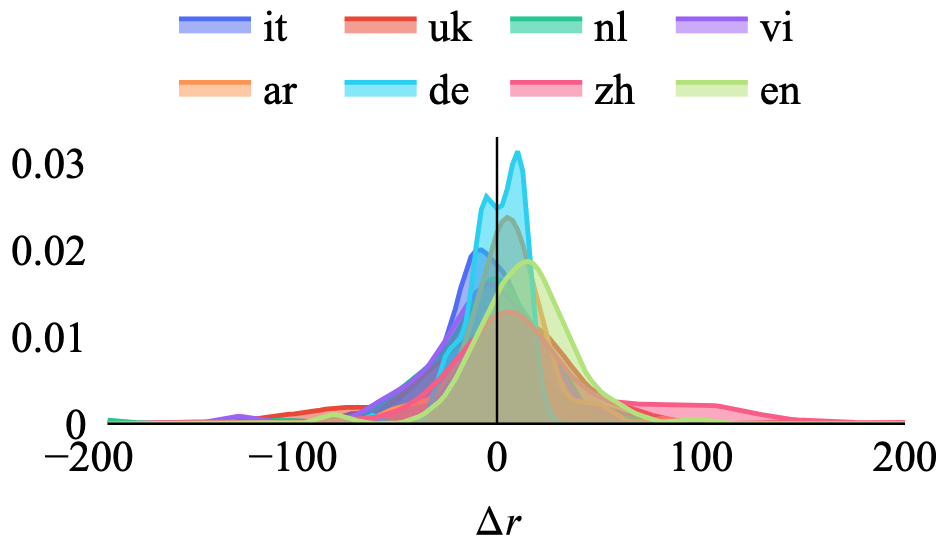}
        \caption{DPO}
        \label{fig:reward_diff_dpo}
    \end{subfigure}
    \hfill
    \begin{subfigure}[b]{0.45\textwidth}
        \centering
        \includegraphics[width=\textwidth]{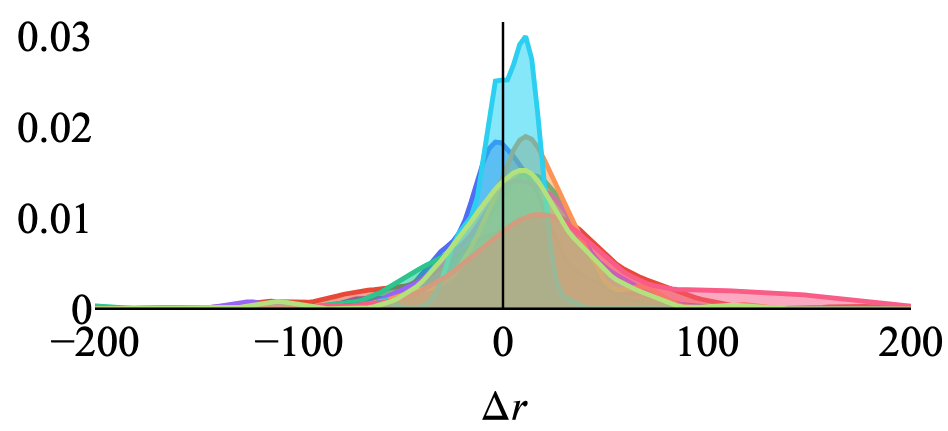}
        \caption{CAPO}
        \label{fig:reward_diff_capo}
    \end{subfigure}
    \caption{Density distribution of reward differences ($\Delta r$) across languages after DPO (top) and after applying our proposed confidence-aware preference optimization - CAPO (bottom). CAPO shifts the distributions towards higher $\Delta r$ values, indicating improved separation between preferred and dispreferred responses.} 
    \label{fig:reward_diff_dpo_capo}
\end{figure}

\begin{figure*}[!t]
    \centering
    \includegraphics[width=1.0\textwidth]{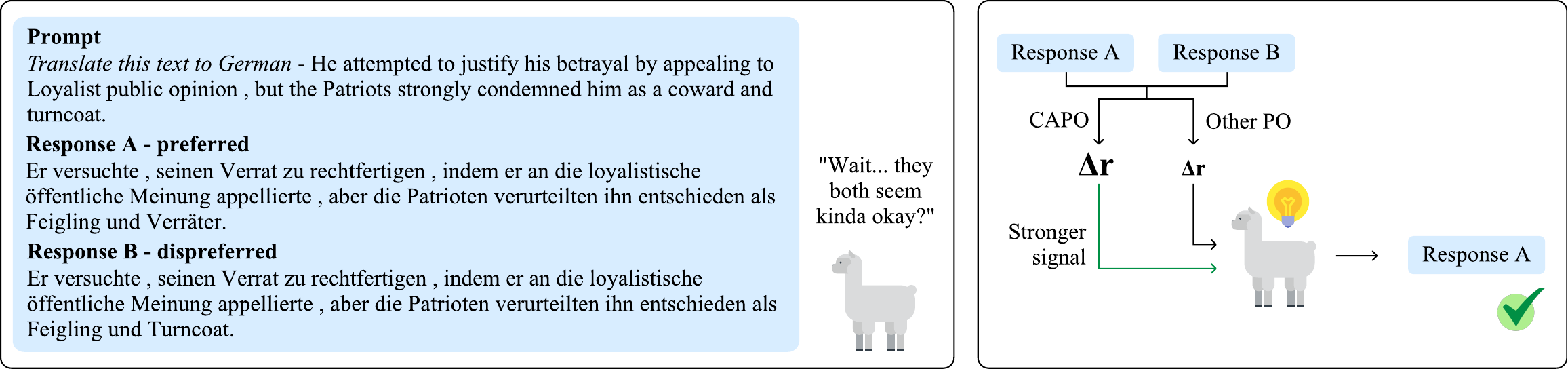}
    \caption{An example of a preference pair where both responses appear similarly plausible. CAPO leverages \texttt{RRM} to interpret such cases and provide a more informative learning signal. {In other words, \texttt{RRM} boosts confidence in favor of the preferred response.}} 
    \label{fig:capo_flowchart}
\end{figure*}


Multilingual alignment \cite{lai2023okapi, wu-etal-2024-reuse, dang-etal-2024-rlhf} increasingly adapts techniques such as PPO \cite{schulman2017proximal}, RLHF \cite{christiano2017deep}, and especially DPO \cite{rafailov2023direct}. However, existing methods still face challenges: DPO treats all preference margins equally \cite{yang2025language}, RLOO relies on reward models, KTO \cite{kto2024} collapses rich feedback to binary labels, and DPL \cite{nath-etal-2025-dpl} uses fixed thresholds that may not generalize. Recent task-specific approaches continue to improve multilingual translation, reasoning, and safety \cite{Xu_contrastive_2024_icml, xu2024x_alma, she-etal-2024-mapo, aakanksha-etal-2024-multilingual}.




We thus propose CAPO\footnote{The code is publicly available at \url{https://github.com/PortNLP/CAPO}.}, Confidence Aware Preference Optimization, a simple yet effective enhancement to DPO that incorporates a Relative Reward Margin (\texttt{RRM}) into the optimization objective. Unlike prior methods that treat all preference pairs equally, our approach dynamically adjusts the loss based on the relative difference in reward scores, allowing the model to calibrate its confidence during training. 

As shown in Figure~\ref{fig:reward_diff_dpo_capo}, this leads to a shift in the reward difference distributions towards the positive side across multiple languages. This is particularly important in multilingual settings, where language-specific inconsistencies and reward model uncertainty can make preference data noisier or less separable. Figure~\ref{fig:capo_flowchart} illustrates how CAPO leverages \texttt{RRM} to improve ambiguous cases by adjusting the learning signal. By reweighting low-confidence or ambiguous examples and emphasizing clearer preference signals, \texttt{RRM} mitigates the risk of overfitting to hard multilingual cases and improves alignment robustness.

{Our key contributions are as follows:}
\begin{itemize}
    \item We propose CAPO, a new PO technique designed for multilingual alignment. CAPO enhances DPO by dynamically adjusting loss based on confidence in preference pairs and also without requiring a reference policy.
    \item We demonstrate CAPO's effectiveness across three multilingual benchmarks, consistently outperforming relevant baselines. 
\end{itemize}



%% file: latex/02_method.tex
\section{CAPO: Confidence Aware Preference Optimization}

\subsection{Limitations of uniform weighting in DPO}  
Given a prompt $x$ along with a preferred (winner) response $y_w$ and a dispreferred (loser) response $y_l$, DPO optimizes the model by directly maximizing the likelihood difference between these responses. Formally, DPO optimizes the reward difference as:

\[
\Delta r = \beta \log \frac{\pi(y_w \mid x)}{\pi_{\text{ref}}(y_w \mid x)} 
- \beta \log \frac{\pi(y_l \mid x)}{\pi_{\text{ref}}(y_l \mid x)}.
\]

It encourages the model to favor the preferred response \( y_w \) over the dispreferred one \( y_l \). While the loss dynamically adjusts its gradient based on the size of \( \Delta r \), it does not distinguish whether the comparison is between two high-quality outputs or two low-quality, poorly aligned ones; it does so only in terms of absolute difference. It does not account for the \textit{relative scale} of rewards across examples or languages. For instance, two preference pairs might yield the same reward difference, $\Delta r = r(y_w \mid x) - r(y_l \mid x) = 1$, but in one case the underlying rewards could be $r = [5, 4]$ and in another $r = [1.2, 0.2]$. 
{The relative preference is much stronger in the later (1.2/0.2 > 5/4). This shows that absolute rewards fail to capture the strength of preference well when rewards are large.}
DPO treats both cases identically. This can become problematic in multilingual alignment, where reward distributions can vary significantly across languages. 
{Empirically, we show that scaling the loss by RRM improves preference signal.}

{In a multilingual setting, tokenization rate varies from language to language. For example - for a similar sample in \textit{en} and \textit{ne}, the reward difference ($\Delta r$) between winning and losing will be bigger for \textit{en} and smaller for \textit{ne}. The bigger reward difference for \textit{en} means small penalty and vice versa. DPO gives a different reward signal in this case, which can unfairly bias optimization toward languages with shorter tokenizations or higher per-token log-probs. Whereas in CAPO, \texttt{RRM} adjusts on a per-example basis to account for these tokenization rate differences. Instead of treating all reward differences uniformly, CAPO scales the loss by the confidence ($\sim$ \texttt{RRM}) derived from the ratio of preferred to dispreferred log-probabilities. \texttt{RRM} is bigger for \textit{ne} which means penalty close to \textit{en}. This ensures that languages with inherently different tokenization structures receive reward signals that better reflect true preference strength rather than tokenization-induced disparities.}

\begin{figure}[!t]
    \centering
    \includegraphics[trim={0 1cm 0 0},clip,width=0.5\textwidth]{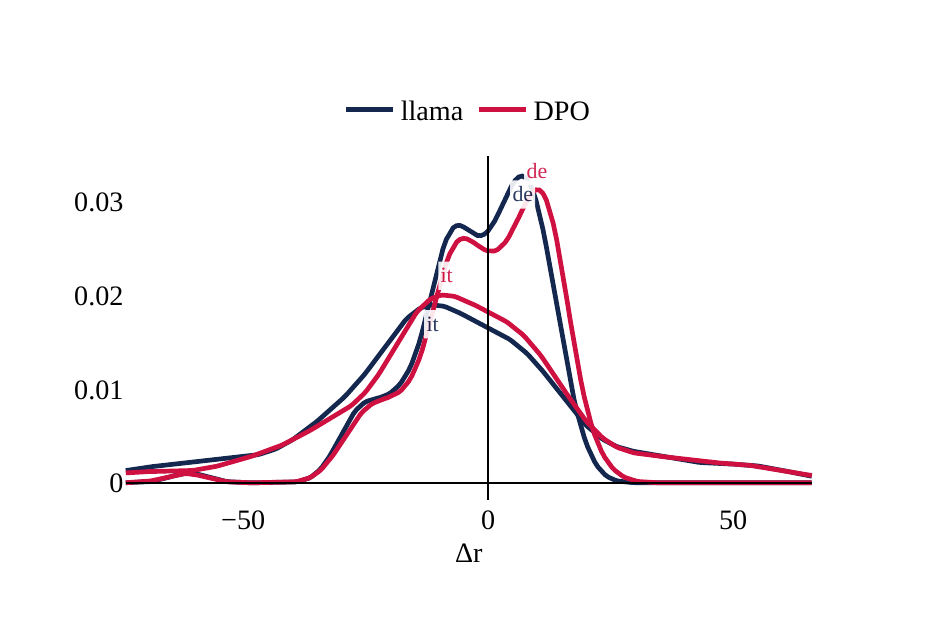}
    \caption{Shift in reward signal across \textit{it} and \textit{de} under llama and DPO. Although there is a shift of reward difference towards the right, there are a lot of samples for which the difference is negative.} 
    \label{fig:reward_diff_dpo}
\end{figure}

Consider an example shown in Figure~\ref{fig:reward_diff_dpo}. While DPO improves the reward margin in languages like \textit{it} and \textit{de}, it still produces a substantial number of preference pairs with negative $\Delta r$, indicating that the model fails to consistently prioritize the intended winning response. This suggests that the same reward signal can have very different implications depending on the language and context. Without a mechanism to calibrate these signals—such as through relative scaling—the model may overreact to weak signals or underreact to strong ones, leading to misaligned or unstable updates. A more robust alignment objective must therefore account not just for the direction of the preference, but also for how decisively that preference is expressed in each sample.

\subsection{Relative Reward Margin}
\label{sec:rrm}
To improve multilingual preference optimization, we propose augmenting the standard DPO framework with a \textit{Relative Reward Margin} (\texttt{RRM}). Given a prompt $x$ with preferred and dispreferred responses $y_w$ and $y_l$, respectively, and a policy model $\pi$ with temperature $\beta$, the log-likelihood terms are:

{\small
\begin{align}
\log \pi(y_w \mid x) &= \text{preferred log-likelihood}, \\
\log \pi(y_l \mid x) &= \text{dispreferred log-likelihood}.
\end{align}
}
Standard DPO is based on the Bradley-Terry (BT) objective, which models the probability of preferring $y_w$ over $y_l$ as a sigmoid function of their log-likelihood difference:

{\small
\begin{equation}
\mathcal{L}_{\text{BT}} = -\log \sigma\left( \beta \left[ \log \pi(y_w \mid x) - \log \pi(y_l \mid x) \right] \right).
\end{equation}
}
We incorporate \texttt{RRM} as a confidence-aware adjustment to this objective by scaling the log-ratio with a dynamic margin term, allowing the model to focus more on examples with clearer preference separation. The resulting modified objective $\mathcal{L}_{\text{CAPO}}$ encourages the policy to increase the relative margin between preferred and dispreferred responses.

{\small
\begin{equation}
\begin{aligned}
\mathcal{L}_{\text{CAPO}}(\pi) = -\mathbb{E}_{(x,y_w,y_l)} \Big[
& \log \sigma\big(\beta(\log \pi(y_w \mid x) \\
&\quad - \log \pi(y_l \mid x)) \\
&\quad + \alpha \cdot \underbrace{\tfrac{\log \pi(y_w \mid x)}{\log \pi(y_l \mid x)}}_{\texttt{RRM}}\big)\Big]
\end{aligned}
\end{equation}
}






\noindent where $\alpha$ is a tunable hyperparameter which adjusts the weight given to \texttt{RRM}. A higher $\alpha$ increases the emphasis on examples with a larger margin between preferred and dispreferred responses, while a lower $\alpha$ reduces this emphasis.  Following \citet{meng2024simpo}, we eliminate the need for a reference model in our objective.
{We omit SimPO’s length normalization since it is problematic in multilingual settings: tokenized lengths vary across languages, leading to unfair penalties for longer-tokenized languages \cite{tsvetkov2024information}.}

The term \texttt{RRM} represents a ratio between preferred and dispreferred rewards and serves to adaptively modulate the loss based on the model’s relative confidence in the given pair. When the preferred reward substantially exceeds the dispreferred one, \texttt{RRM} increases and the overall loss diminishes—signaling that the model is already well-aligned in this example. In contrast, when the reward margin is small, \texttt{RRM} downweights the learning signal to avoid overfitting on uncertain or noisy preference pairs. In other words, it dynamically calibrates the optimization signal based on the model's confidence in each preference pair, particularly beneficial in multilingual alignment, where preferences are harder to model due to linguistic variability. By emphasizing confident updates and softening gradients on ambiguous cases, our method encourages more stable convergence and robust generalization across diverse languages.

\begin{table*}
\centering
\small
\setlength{\tabcolsep}{10pt}
\begin{tabular}{@{} p{4.4cm} p{8.7cm} @{}}
\toprule
\multicolumn{1}{c}{\textbf{Method}} & \multicolumn{1}{c}{\textbf{Objective}} \\
\midrule
DPO~\cite{rafailov2023direct} &
$-\log \sigma \left( \beta \log \frac{\pi_\theta(y_w|x)}{\pi_{\text{ref}}(y_w|x)} - \beta \log \frac{\pi_\theta(y_l|x)}{\pi_{\text{ref}}(y_l|x)}\right)$ \\
\midrule
SimPO~\cite{meng2024simposimplepreferenceoptimization} &
$-\log \sigma \left( \frac{\beta}{|y_w|} \log \pi_\theta(y_w|x) - \frac{\beta}{|y_l|} \log \pi_\theta(y_l|x) - \gamma \right)$ \\
\midrule
DPONLL \cite{yang2025language} & $-\log \sigma \left( \beta \log \frac{\pi_\theta(y_w \mid x)}{\pi_{\text{ref}}(y_w \mid x)} - \beta \log \frac{\pi_\theta(y_l \mid x)}{\pi_{\text{ref}}(y_l \mid x)} \right) + \text{NLL}$ \\
\midrule
\midrule
\textbf{CAPO} & $-\log \sigma\left( \beta \log \pi(y_w \mid x) - \beta \log \pi(y_l \mid x) + \alpha \cdot \frac{\log \pi(y_w \mid x)}{\log \pi(y_l \mid x)} \right)$ \\
\bottomrule
\end{tabular}
\caption{Various preference optimization objectives given preference data $\mathcal{D} = (x, y_w, y_l)$, where $x$ is an input and $y_w$ and $y_l$ are the winning and losing responses.} 
\label{tab:all_dpo_methods}
\end{table*}

\begin{table*}[!t]
\centering
\small
\begin{tabular}{p{1cm} p{3.9cm} p{3.9cm} p{3.9cm} p{1cm}}
\toprule
\multirow{2}{*}{\makecell{\textbf{Lang.}\\\textbf{Direction}}}
  & \colorbox{blue!20}{\texttt{prompt}}
  & \colorbox{red!20}{\texttt{dispreferred}}
  & \colorbox{green!20}{\texttt{preferred}}
  & \multirow{2}{*}{\textbf{Dataset}} \\
 & \textbf{(Source Text)} \rule{0pt}{2.6ex}
 & \textbf{(Machine translation)} \rule{0pt}{2.6ex}
 & \textbf{(MTPE text)} \rule{0pt}{2.6ex}
 &  \\
\midrule
en-tr &People may not anticipate that patience and understanding are also necessary for travellers returning home. &İnsanlar bu sabır tahmin edemeyeceğiniz ve anlayış da eve dönen yolcular için gereklidir. &İnsanlar, ülkesine dönenlere de sabır ve anlayış gösterilmesi gerektiğini tahmin edemeyebilir. &DIVEMT \\
en-de &He also begins an affair with Veronica Harrington, who bails him out. &Er beginnt auch eine Affäre mit Veronica Harrington, die ihn rettet. &Er beginnt auch eine Affäre mit Veronica Harrington, die ihn rettet. &MLQE-PE \\
\bottomrule
\end{tabular}
\caption{A sample each from the MTPE datasets.} 
\label{tab:mtpe_samples}
\end{table*}

\section{Experimental Setup}


\subsection{Models and Implementation} We use \texttt{Llama-3.1-8B-Instruct} (llama) as the base model and apply parameter-efficient fine-tuning using LoRA \cite{hu2022lora} (additional LoRA settings can be found in Appendix \ref{sec:app_implementation}).
The chosen and rejected samples are fed to the model in the form of a dialogue, where the examples are formatted using a custom Zephyr-style chat template \cite{tunstall2024zephyr}. We conduct hyper parameter search to set the value of $\alpha$, which is set to 2.0 based on maximum evaluation accuracy during training (more details in Section~\S\ref{subsec:alpha_change}).

\subsection{Baselines} Consistent with prior work \cite{dang-etal-2024-rlhf}, as our baselines, we consider base llama along with tuned versions of  llama with DPO  \cite{rafailov2023direct}, {SimPO \cite{meng2024simpo}, and DPONLL \cite{yang2025language}}. For DPO setup, the reference policy is identical to the policy being optimized. We selected DPO as a relevant baseline over SFT, given prior studies demonstrating its superior performance \cite{wu-etal-2024-reuse, yang2025language, yang2025implicit}, and over RLHF due to its greater efficiency {\cite{rafailov2023direct}. We use the same LoRA setting for finetuning across all the objectives compared. Table~\ref{tab:all_dpo_methods} presents the optimization objectives evaluated in this work.

\subsection{Preference Dataset for Training} 


Multiple methods for creating multilingual preference datasets have been explored \citep{dang-etal-2024-rlhf, she-etal-2024-mapo, yang2025language, aakanksha-etal-2024-multilingual}; however, none of these datasets have been publicly released. As such, inspired by \citet{berger-etal-2024-post}, where they used Machine Translated Post Edited (MTPE) data for preference alignment on machine translation, we use MTPE data as our preference data. MTPE data provides a natural, human-grounded preference signal: the post-edited output reflects human judgments of correctness and fluency, while the corresponding raw machine translation often contains errors or stylistic flaws. This eliminates the need for synthetic preference labels and introduces realistic, linguistically diverse “hard examples”.


Specifically, we repurpose two MTPE datasets—\texttt{DIVEMT} \citep{sarti-etal-2022-divemt} and \texttt{MLQE-PE} \citep{fomicheva-etal-2022-mlqe}—into preference datasets, although they were not originally designed/used for this purpose. Together, they cover eight language directions with English as the source: \textit{en-ar}, \textit{en-it}, \textit{en-nl}, \textit{en-tr}, \textit{en-uk}, \textit{en-vi} (\texttt{DIVEMT}) and \textit{en-de}, \textit{en-zh} (\texttt{MLQE-PE}).


As shown in Table~\ref{tab:mtpe_samples}, each sample includes a source sentence, a machine translated output, and a post-edited version, where the source sentence is the \colorbox{blue!20}{\texttt{prompt}}, the post-edited version is the \colorbox{green!20}{\texttt{preferred}}, and the machine translation is the \colorbox{red!20}{\texttt{dispreferred}}. The prompt part is formatted using the following prompt template.
\begin{quote}
\small
\texttt{Translate this text from \{src\_lang\} to \{tgt\_lang\}:\\\\
\{src\_lang\}: \{src\_sent\}\\
\{tgt\_lang\}:}
\end{quote}
We filter the dataset to include only samples where the text length exceeds 50 characters, to make sure the model receives sufficient context.  Data is balanced per translation direction, with 100 samples per direction. This gives us 800 samples for training and 800 for evaluation. All related analysis are done on the evaluation set. We perform validation on a held out 800 samples built in the same way as the training dataset for hyperparameter tuning.



\subsection{Evaluation Benchmarks and Metrics} We consider three widely used multilingual benchmarks for evaluation (see implementation details in Appendix~\ref{sec:app_implementation}).

\paragraph{\em Multilingual MT-Bench} \cite{zheng2023judging} {evaluates  the capabilities and alignment of language models through two-turn dialogue prompts, covering open-ended tasks like writing, reasoning, and math. Models generate responses for each turn based on prompts provided by the benchmark. GPT-4-Turbo serves as the judge and assigns a score from 1 to 10 for each turn based on overall quality, including helpfulness, relevance, and fluency. We report the average of the two turns. We test on seven languages: the seen languages \textit{[en, zh, it, and de]}, and the unseen languages \textit{[fr, es, and jp]}, with 80 samples per language. 
\paragraph{\em XLSum} \cite{hasan-etal-2021-xl} is a multilingual abstractive summarization dataset consisting of BBC news articles across a wide range of languages. Following prior work on multilingual preference optimization \cite{dang-etal-2024-rlhf}, we evaluate on {50 samples from each of} six \textit{seen} languages: \textit{[en, zh, vi, ar, uk, tr]}, and three \textit{unseen} languages: \textit{[fr, ja, es]}. {We report win rates with GPT4o as the judge model (and Rouge-L scores under Appendix~\ref{sec:benchmarks})}.

\paragraph{\em M-IFEval} \cite{dussolle-etal-2025-ifeval} is a multilingual instruction‑following benchmark that covers 4 languages: one \textit{seen} - \textit{[en]}, and three \textit{unseen} - \textit{[fr, ja, es]}. It evaluates multilingual instruction-following by models using objective checks like string matching and rule-based evaluation rather than relying on the LLM-as-judge model. We report both strict (exact matches) and loose (acceptable variations) evaluation metrics.

\begin{figure}[!t]
    \centering
    \includegraphics[trim={0 0.65cm 0 1cm},clip,width=0.5\textwidth]{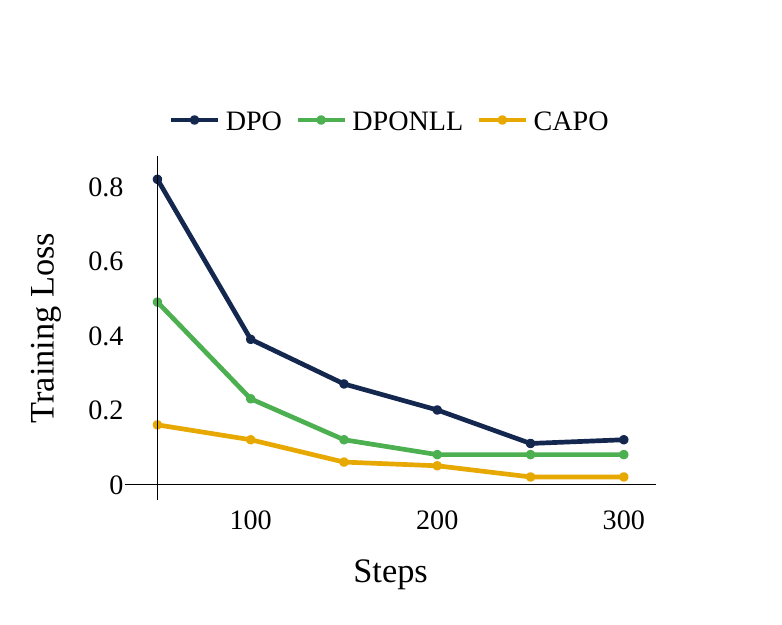}
    \caption{Training Loss vs. Steps for DPO and CAPO: CAPO demonstrates  improved stability and convergence over DPO and DPONLL.}  
    \label{fig:training_loss}
\end{figure}




%% file: latex/03_results.tex
\section{Results and Analysis}

\begin{figure}[!t] 
    \centering
    \includegraphics[trim={0.5cm 1.5cm 0.5cm 1cm},clip,width=0.55\textwidth]{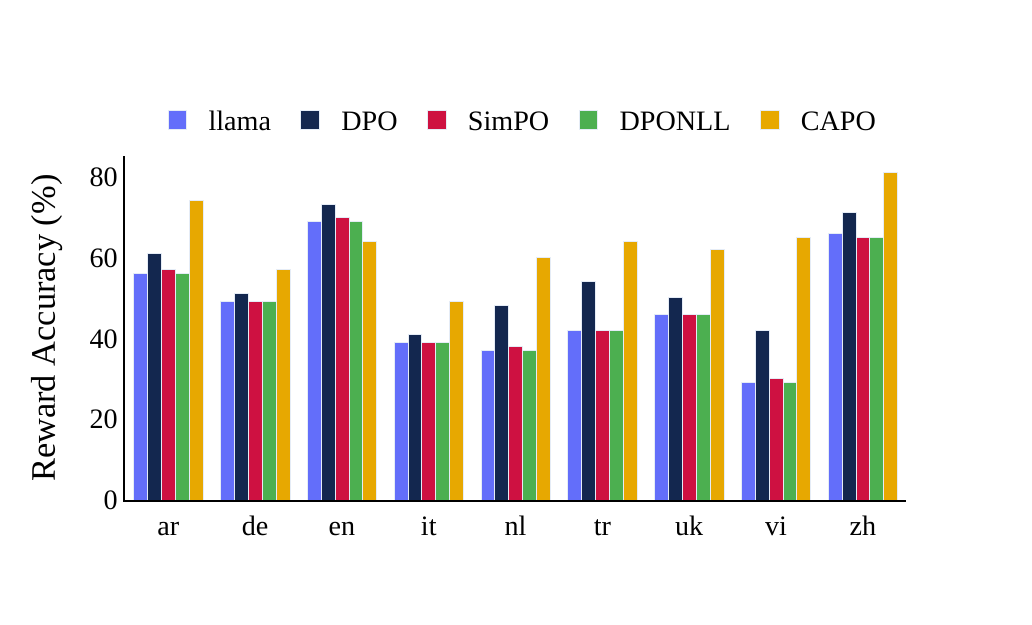}
    \caption{Comparison of reward accuracy between DPO and CAPO on the validation data. {CAPO shows improved accuracy across all languages except in \textit{en}.}} 
    \label{fig:reward_acc_perlang}
\end{figure}

\begin{figure*}[!t]
    \centering
    \includegraphics[trim={0 1.5cm 0 1.5cm},clip,width=\textwidth]{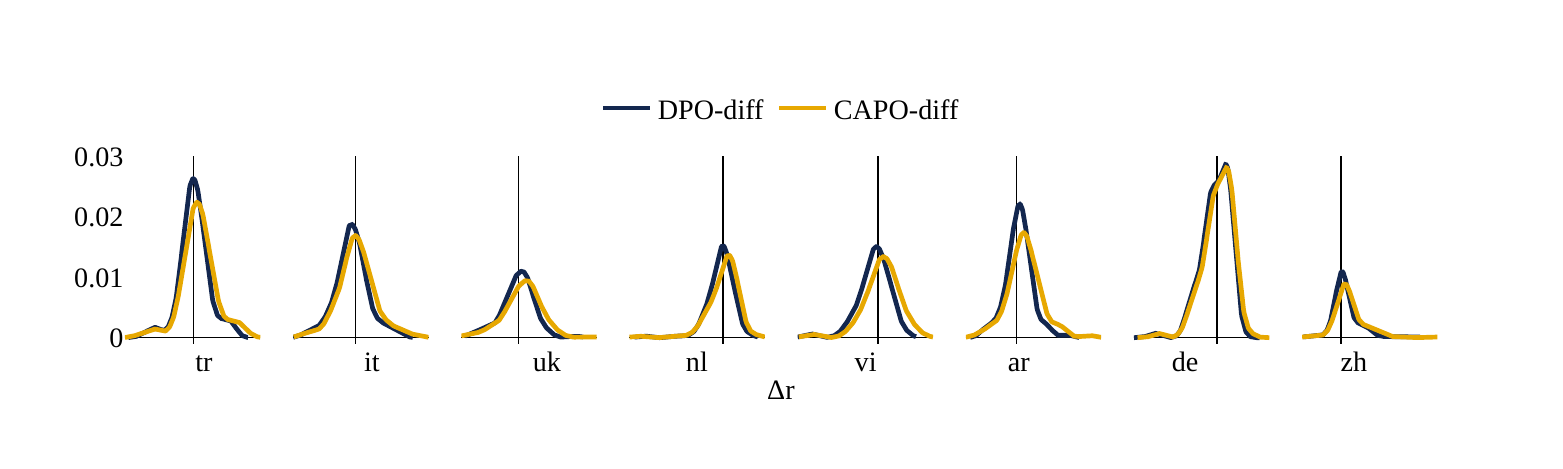}
    \caption{Reward difference distribution between DPO vs CAPO per language.  {CAPO consistently shifts the distribution toward higher reward differences.}}
    \label{fig:reward_shift_perlang}
\end{figure*}

\begin{figure}[!t]
    \centering
    \includegraphics[trim={0 1.3cm 0 0},clip,width=0.35\textwidth]{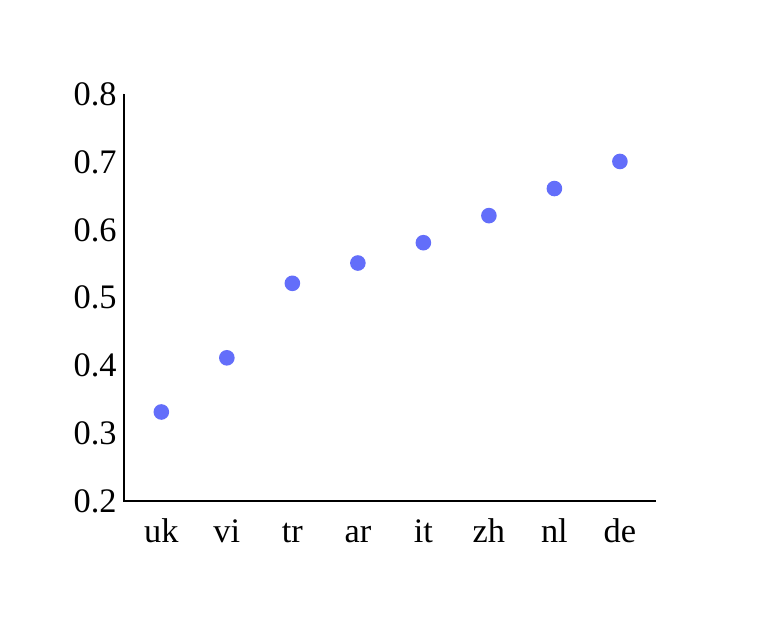}
    \caption{BLEU score between the preferred and dispreferred samples in the training data per language.}
    \label{fig:bert_mtpe} 
\end{figure}

\begin{figure*}[!t]
    \centering
    \includegraphics[width=1.0\textwidth]{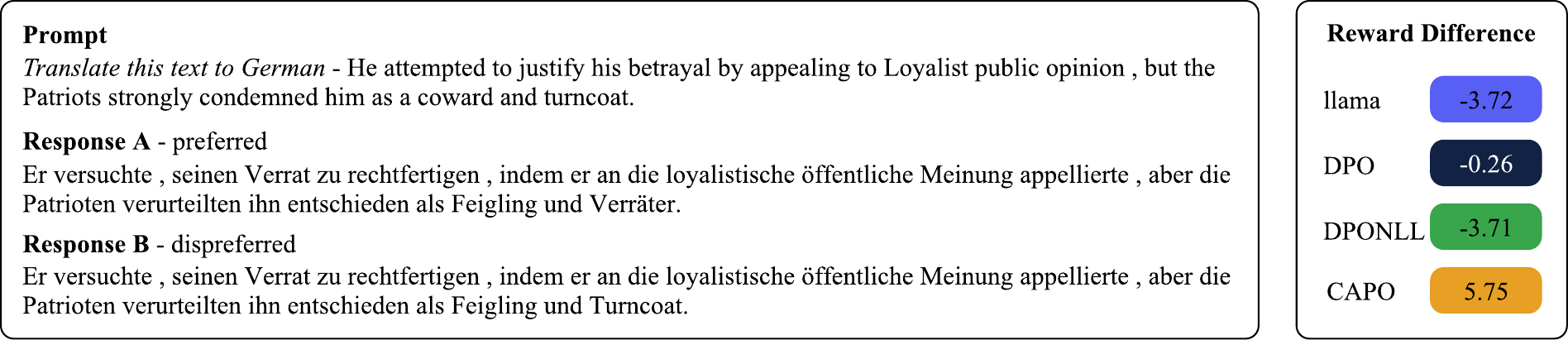}
    \caption{The reward difference shows how well each model distinguishes the responses. CAPO increases the gap between preferred and dispreferred responses the most.}  
    \label{fig:reward_gaps}
\end{figure*}

{The following subsections present key findings from our experiments with CAPO. We show that it improves training stability, sharpens reward signals, and increases the gap between preferred and dispreferred outputs. We also examine the effect of varying the $\alpha$ parameter. Finally, we report results on multilingual benchmarks.}

\subsection{CAPO makes training more stable.} 
We compare training curves for CAPO against two baselines: DPO and DPONLL \cite{yang2025language} in Figure~\ref{fig:training_loss}. We observe clear differences in stability and convergence. DPO steadily improves reward but shows some fluctuations in later epochs. DPONLL is more stable, with smoother and more consistent reward growth. This stability comes from the added negative log-likelihood reweighting. CAPO shows the most stable and monotonic reward increase. It learns faster in early stages and then levels off gradually. This suggests that CAPO’s confidence-aware design speeds up early training and supports stable alignment later. 

We further investigate the lower training loss of CAPO compared to DPO and DPONLL. While DPO uses a contrastive term based on the log-sigmoid of log-probability differences, and DPONLL adds an unbounded negative log-likelihood (NLL) term that heavily penalizes low-probability preferred responses, CAPO avoids such penalties. Instead, CAPO combines the bounded ratio-based component (\texttt{RRM}) that contributes a smaller signal than NLL term. Since it adds only a mild adjustment (compared to NLL) on top of the reward difference, the overall loss remains lower. This makes CAPO's loss lower from the start.


\subsection{The reward signal to the model gets better with CAPO.}
\label{subsec:capo_vs_po}
Reward accuracy refers to how often a reward model assigns a higher score to the response that humans prefer in a pairwise comparison (more details under Appendix \ref{sec:reward_cal}). Figure~\ref{fig:reward_acc_perlang} shows results across eight languages and 5 settings. On average, CAPO significantly outperforms DPO by 16\% and DPONLL by 33\%. It achieves the highest reward accuracy in all languages except English \textit{en}. CAPO’s poor performance in \textit{en} may be due to using EN $\rightarrow$ other languages MTPE data, which might have introduced learned preferences that are subtly different from native English data. Vietnamese \textit{vi} enjoys the most gain just like in the standard DPO. These results suggest that CAPO not only improves training stability but also leads to better reward modeling aligned with human intent. We attribute this to \texttt{RRM}, which helped deliver more reliable reward signals during optimization.

\subsection{CAPO leads to increase in reward difference.}

Figure~\ref{fig:reward_shift_perlang} shows the distribution of reward differences between preferred and dispreferred responses under DPO and CAPO across eight languages. The reward difference reflects how confidently the model separates the preferred response from the rejected one. 
Compared to DPO (where the curves are more narrowly peaked around zero), CAPO's distributions often show a broader spread with a greater mass concentrated on the right side of zero on all languages. This indicates that under CAPO, the reward model assigns significantly higher scores to preferred responses more frequently, leading to stronger reward separation. 

We conduct further analysis. Languages like \textit{vi}, \textit{ar}, and \textit{uk} show larger shift towards the right which can be credited to the lower BLEU scores (i.e.\ lower similarity) in these languages (see Figure~\ref{fig:bert_mtpe}) between preferred and dispreferred samples. This hints that languages exhibit clearer preference signals benefit more from CAPO. Supporting this, we observe an inverse correlation ($r = -0.47$) between BLEU scores and the CAPO-vs-DPO reward shift across languages, computed by correlating BLEU scores with the difference in KDE-weighted means of reward distributions. 

For instance, Figure~\ref{fig:reward_gaps} shows an example where the initial reward gap between the preferred (Response A) and dispreferred (Response B) is negative, with llama assigning -3.72. DPO slightly improves this to -0.26, while DPONLL remains similar at -3.71. CAPO significantly increases the reward gap to 5.75, showing a much stronger distinction in favor of the preferred response.


\begin{figure}[!t]
    \centering
    \includegraphics[trim={0 0.5cm 0 0},clip,width=0.4\textwidth]{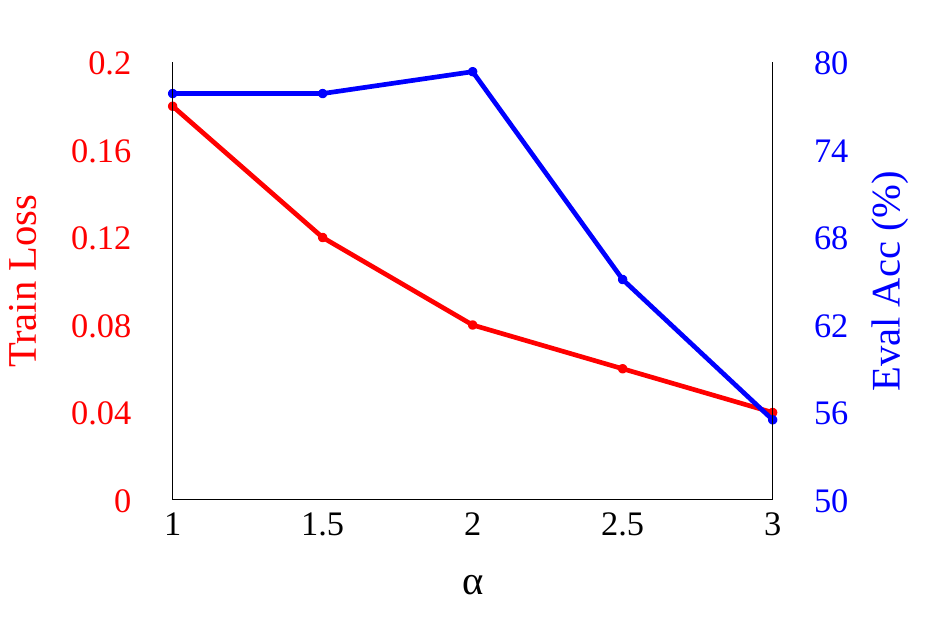}
    \caption{Impact of $\alpha$ on training loss and evaluation accuracy.} 
    
    \label{fig:alpha_change}
\end{figure}

\subsection{Impact of varying $\alpha$.}
\label{subsec:alpha_change}
We investigate the impact of the reweighting coefficient $\alpha$ in the $\mathcal{L}_{\text{CAPO}}$ objective, where $\alpha$ scales the log-ratio term that encourages stronger separation between preferred and rejected responses. As shown in Figure~\ref{fig:alpha_change}, increasing $\alpha$ consistently reduces training loss, indicating improved optimization. However, evaluation accuracy exhibits a tradeoff: it increases up to $\alpha = 2.0$ but drops sharply beyond that point. This suggests that while moderate reweighting strengthens preference alignment, excessive weighting of the \texttt{RRM} term leads to over-reweighting and harms generalization. Overall, the results highlight a critical tradeoff between training loss minimization and evaluation performance when tuning $\alpha$.

\subsection{Multilingual Benchmarks}
Next, we compare CAPO with the strongest baseline DPO in the following downstream benchmarks.

\begin{table}[!t]
\centering
\small
\setlength{\tabcolsep}{14pt}
\begin{tabular}{p{0.4cm} p{0.4cm} p{0.4cm} p{0.4cm} p{0.4cm}}
\toprule
\textbf{Lang.} & \textbf{llama} & \textbf{DPO} & \textbf{SimPO} & \textbf{CAPO} \\
\midrule
en  & 7.83 & \textbf{8.12} & 6.49 & 8.08 \\
de  & \textbf{7.43} & 7.01 & 7.25 & 7.07 \\
it  & 6.80 & 7.16 & 6.87 & \textbf{7.39} \\
zh  & 6.55 & 6.60 & 6.87 & \textbf{7.62} \\
fr  & 7.01 & 6.93 & 6.72 & \textbf{7.40} \\
es  & 6.73 & 7.07 & \textbf{7.72} & 7.09 \\
ja  & 6.85 & 6.71 & 6.75 & \textbf{7.18} \\
\midrule
Avg & 7.03 & 7.09 & 6.95 & \textbf{7.40} \\
\bottomrule
\end{tabular}
\caption{Results on multilingual MT-Bench benchmark.} 
\label{tab:mt_bench}
\end{table}

\begin{figure}[!t]
    \centering
    \begin{subfigure}[b]{0.5\textwidth}
        \centering
        \includegraphics[trim={0 1.5cm 0 1.5cm},clip,width=\textwidth]{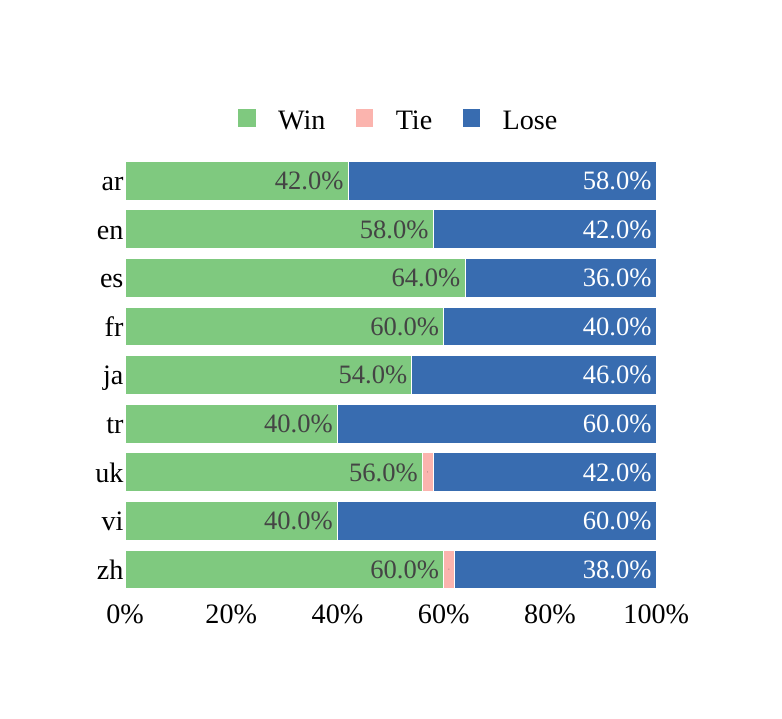}
        \caption{Win rate of CAPO against llama.}
        \label{fig:xlsum_winrate_capo_llama}
    \end{subfigure}
    \hfill
    \begin{subfigure}[b]{0.5\textwidth}
        \centering
        \includegraphics[trim={0 1.5cm 0 1.5cm},clip,width=\textwidth]{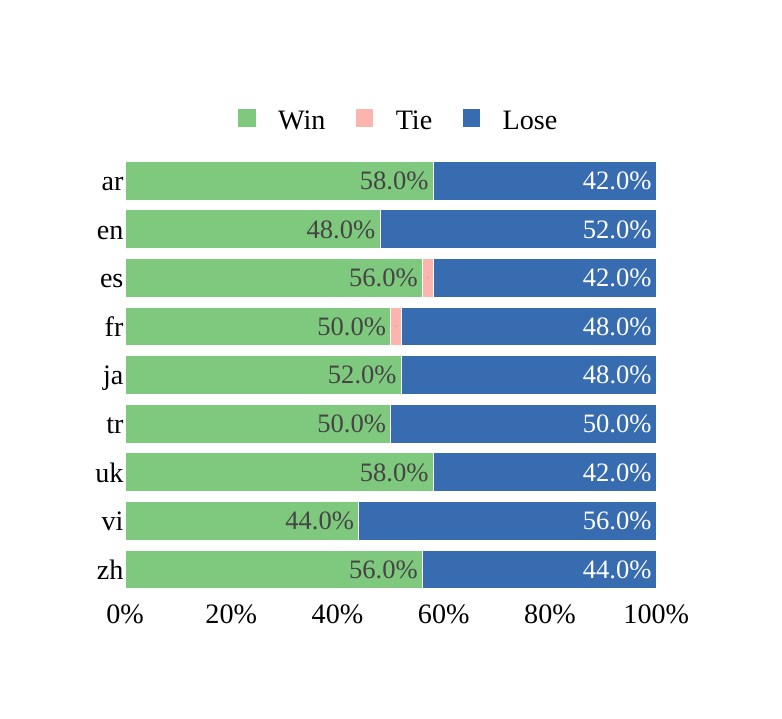}
        \caption{Win rate of CAPO against DPO.}
        \label{fig:xlsum_winrate_capo_dpo}
    \end{subfigure}
    \caption{Comparison of CAPO's win rates against llama and DPO for XLSum summaries. {CAPO shows improved ability to generate better summaries across languages compared to the baselines.} Standard deviations across runs are in Appendix \ref{sec:app_results}.} 
    \label{fig:xlsum_winrate_comparison}
\end{figure}

\begin{table*}[!t]
\centering
\footnotesize
\begin{tabular}{p{4.4cm}p{3.4cm}p{3.4cm}p{3.4cm}}
\toprule
\textbf{Text} &\textbf{llama} &\textbf{DPO} &\textbf{CAPO}\\
\midrule
The application is designed to stop using the computer's central processing unit (CPU) when it is left to run in the background. It means users can receive calls or instant messages without the risk of exhausting their laptop battery. The move may help Skype [...] &Microsoft has released a new version of Skype that can pause and resume activity without using the computer's central processing unit, allowing users to receive calls or messages without draining their laptop battery. &The software update for Skype allows the program to run in the background without consuming the computer's CPU power, thus saving battery life for laptops and tablets. &The new version of Skype will conserve laptop battery life by automatically suspending activity when the application is left to run in the background.\\
\bottomrule
\end{tabular}
\caption{An example of the summaries generated by different models and their head to head comparison against CAPO. Here, CAPO's answer is selected by the judge (GPT4o) both the times.} 
\label{tab:xlsum_summaries}
\end{table*}


\paragraph{Multi-turn Conversation and Instruction Following (Multilingual MT-Bench)}
Table~\ref{tab:mt_bench} presents results on the multilingual MT-Bench benchmark across seven languages. CAPO achieves the highest average score (7.40), outperforming  DPO (7.09), SimPO (6.95), and the base llama model (7.03). Among the seen languages (\textit{en}, \textit{de}, \textit{it}, \textit{zh}), CAPO performs competitively, with notable improvements in \textit{it} and \textit{zh}, suggesting better generalization in training-aligned languages. For unseen languages (\textit{fr}, \textit{es}, \textit{ja}), CAPO also shows consistent gains over other models except for \textit{es}. This indicates that the confidence-aware objective used in CAPO improves not only in-domain alignment but also enhances cross-lingual generalization to languages not observed during training.

\paragraph{Summarization (XLSum)}
Figure~\ref{fig:xlsum_winrate_comparison} shows the pairwise win rates of CAPO compared to the base llama model (Figure~\ref{fig:xlsum_winrate_capo_llama}) and DPO (Figure~\ref{fig:xlsum_winrate_capo_dpo}) across eight languages. Against llama, CAPO achieves substantial improvements in most languages, winning over 50\% of the time in \textit{en}, \textit{zh}, \textit{uk}, \textit{fr}, \textit{ja}, and especially \textit{es} (64\%). This demonstrates that preference optimization significantly enhances response quality over the base model. When compared to DPO, the margins are narrower but still favorable to CAPO in most cases. CAPO outperforms DPO in six out of eight languages, with strong gains in \textit{ar}, \textit{uk}, \textit{zh}, and \textit{es}, where win rates reach or exceed 56\%. DPO outperforms CAPO is \textit{en} and \textit{vi}, with a 52\% and 56\% win rate respectively. These results indicate that CAPO offers robust multilingual improvements.

We further examined sample summaries generated by llama, DPO, and CAPO in en. Table~\ref{tab:xlsum_summaries} presents one example from each model for the same input text. A judge model compared the CAPO summary against those from llama and DPO. In both cases, the CAPO summary was rated better than the other two.

\begin{table}[!t]
\small
\centering
\begin{tabular}{ccrrr}
\toprule
\textbf{Lang.} & \textbf{Score Type} &\textbf{DPO} &\textbf{CAPO} \\
\midrule
en &Strict &0.46 &\textbf{0.47} \\
 &Loose &0.6 &\textbf{0.63} \\
es &Strict &\textbf{0.56} &0.55 \\
 &Loose &0.68 &\textbf{0.69} \\
fr &Strict &0.46 &\textbf{0.48} \\
 &Loose &0.63 &\textbf{0.64} \\
ja &Strict &0.23 &\textbf{0.29} \\
 &Loose &0.41 &\textbf{0.43} \\
\bottomrule
\end{tabular}
\caption{Comparison between the average scores (strict and loose) of M-IFEval using DPO and CAPO.} 
\label{tab:ifeval_results}
\end{table}

\paragraph{Instruction‑Following (M-IFEval)}
Table \ref{tab:ifeval_results} reports the accuracy of the response for the given prompts. Here again, we can see that CAPO consistently outperforms DPO under both the settings.

%% file: latex/04_related_work.tex
\section{Related Work}

Broadly, multilingual PO research can be categorized into general-purpose methods that align models across multiple languages using shared optimization mechanisms ~\cite{dang-etal-2024-rlhf,yang2025implicit,yang2025language}, and task-specific strategies adapted to domains such as multilingual reasoning, safety alignment, text quality evaluation, and machine translation~\cite{she-etal-2024-mapo,aakanksha-etal-2024-multilingual,Xu_contrastive_2024_icml,xu2024x_alma, mtqeval2025}.


A major focus of recent work has been on cross-lingual alignment—developing techniques to transfer preferences learned in one language (typically \textit{en}) to others. Most prior approaches generate synthetic multilingual preference data via translation. \citet{lai2023okapi} translate both prompts and responses, \citet{dang-etal-2024-rlhf} translate only prompts, and \citet{yang2025language} translate completions to favor dominant language outputs. While these cross-lingual strategies have enabled preference alignment in many languages, they often carry over linguistic artifacts and biases from the source language, which may not reflect the true human preference.

Regarding PO objectives, earlier studies have extended a range of PO techniques originally developed for \textit{en}. \citet{lai2023okapi} and \citet{wu-etal-2024-reuse} employ standard reinforcement learning methods such as Proximal Policy Optimization (PPO) \cite{schulman2017proximal}, with the latter placing emphasis on cross-lingual transfer during optimization. Cal-DPO \cite{xiao2024caldpo} calibrates implicit rewards to fixed targets using a reference model.
{Likewise, SimPO operates in a reference-free setting and applies length normalization, which is problematic in multilingual setting as discussed in Section~\ref{sec:rrm}. CPO \cite{cpo2024} uses a contrastive loss to distinguish preferred from dispreferred responses based on their log-probability differences and, similar to DPO, treats all pairs uniformly. Furthermore, CPO depends on a reference model, adding an additional layer of complexity.} 

Similarly, \citet{dang-etal-2024-rlhf} explore Reinforcement Learning with Optimal Outputs (RLOO) alongside DPO, while \citet{yang2025language} apply DPO in conjunction with NLL of the preferred response. Although RLOO improves upon traditional RLHF approaches \cite{christiano2017deep} and PPO by reducing variance through multi-sample baselines, it still depends on a separately trained reward model, which introduces potential alignment gaps. 
Another PO method, Kahneman-Tversky Optimization \cite{kto2024}, uses a prospect-theoretic objective to model human-like cognitive biases, but relies on binary desirability labels instead of preference pairs. Diverse Preference Learning~\cite{nath-etal-2025-dpl} adjusts loss contributions based on the contrast between preferred and dispreferred samples, amplifying clear preferences and downweighting ambiguous ones. However, it relies on fixed thresholds and global weightings, which do not adapt to the reward dynamics of individual examples.

%% file: latex/05_conlcusion.tex
\section{Conclusion}

This paper introduces CAPO, a novel multilingual alignment objective that utilizes relative reward differences between preferred and dispreferred responses to guide the model's alignment. CAPO avoids reliance on a reward model and bypasses cross-lingual alignment methods that risk introducing translationese. Through empirical results on multilingual benchmarks, CAPO demonstrates significant improvements in aligning model outputs with human preferences across diverse languages, achieving 16–33\% gains in reward accuracy. While we relied on existing MTPE data for training, future work should consider  automatically generating high-quality MTPE data to expand language coverage and better reflect true human preferences.  


%% file: latex/06_appendix.tex
\section*{Appendix}
\label{sec:appendix}

\section{Implementation Details}
\label{sec:app_implementation}

\noindent \textbf{Finetuning}
Following the configuration of \citet{thakkar-etal-2024-deep} on DPO finetuning, we set the LoRA rank to 16, alpha to 32, and apply a dropout rate of 0.05. Likewise, optimization is conducted using the Paged AdamW optimizer, paired with a cosine decay learning rate scheduler. Training is performed using the CAPO objective for 1 epoch, with a batch size of 16 and a learning rate of $1\mathrm{e}{-6}$. We use a temperature parameter of $\beta = 0.1$, consistent with prior work.

\noindent \textbf{MT-bench}
We use the multilingual MT-Bench framework \footnote{https://github.com/lightblue-tech/multilingual-mt-bench} \cite{zheng2023judging}, following the same setup as \citet{yang2025language}. The benchmark extends the original MT-Bench with multilingual support in \textit{en}, \textit{de}, \textit{it}, \textit{zh}, \textit{fr}, \textit{es}, and \textit{ja}, with 80 samples per language. 

The base model evaluated in all experiments is \texttt{Llama 3.2 8B Instruct}. We compare three variants: the untuned base model, a DPO-finetuned model, and our proposed CAPO-finetuned model. All generations are produced using identical decoding settings, with a maximum of 1024 generated tokens. Sampling temperature is set to 0.7 by default. To ensure a fair comparison, the same evaluation pipeline is used across all model variants.

\noindent \textbf{XLSum}
The evaluation prompt used to calculate win rate between two summaries is adapted from \citet{yang2025language}, which is shown in Figure~\ref{fig:xlsum_judge_prompt}. To mitigate order bias, the summaries were randomly shuffled.

\section{Reward Accuracy Calculation}
\label{sec:reward_cal}
As in SimPO, we compute \textit{reward accuracy} directly from the trained model's own log-probabilities, without using a reference model. For each preference pair $(x, y_w, y_l)$, we calculate the model scores as the log-likelihood of each response, following SimPO's formulation. Reward accuracy is then defined as the percentage of cases where the model assigns a higher score to the preferred response $y_w$ than to the dispreferred response $y_l$.

\begin{figure}[htbp]
\begin{tcolorbox}[
  colback=gray!5!white,
  colframe=green!50!black,
  title=XLSum Win Rate Prompt,
  listing only,
  listing options={
    breaklines=true,
    columns=fullflexible,
    keepspaces=true,
    upquote=true
  }
]
Which of the following answers is the best one for given instruction in [LANGUAGE]. A good answer should follow these rules:\\

1) It should be in [LANGUAGE]\\
2) It should answer the request in the instruction\\
3) It should be factually and semantically comprehensible\\
4) It should be grammatically correct and fluent.\\

Instruction: Generate a one sentence summary of the text below in [LANGUAGE].\\

[TEXT]\\

Answer (A): [SUMMARY\_A]\\
Answer (B): [SUMMARY\_B]\\

FIRST provide a one-sentence comparison of the two answers, explaining which you prefer and why.\\
SECOND, on a new line, state only `Answer (A)' or `Answer (B)' or `TIE'.\\

Your response should use the format:\\
Comparison: <one-sentence comparison and explanation>\\
Preferred: <`Answer (A)' or `Answer (B)' or `TIE'>
\end{tcolorbox}
\caption{Prompt used for win rate calculation on XLSum summaries.}
\label{fig:xlsum_judge_prompt}
\end{figure}

\section{Multilingual Benchmarks}
\label{sec:benchmarks}

This section outlines the remaining analyses on the multilingual benchmarks.

\begin{figure}[!t]
    \centering
    \includegraphics[width=0.50\textwidth]{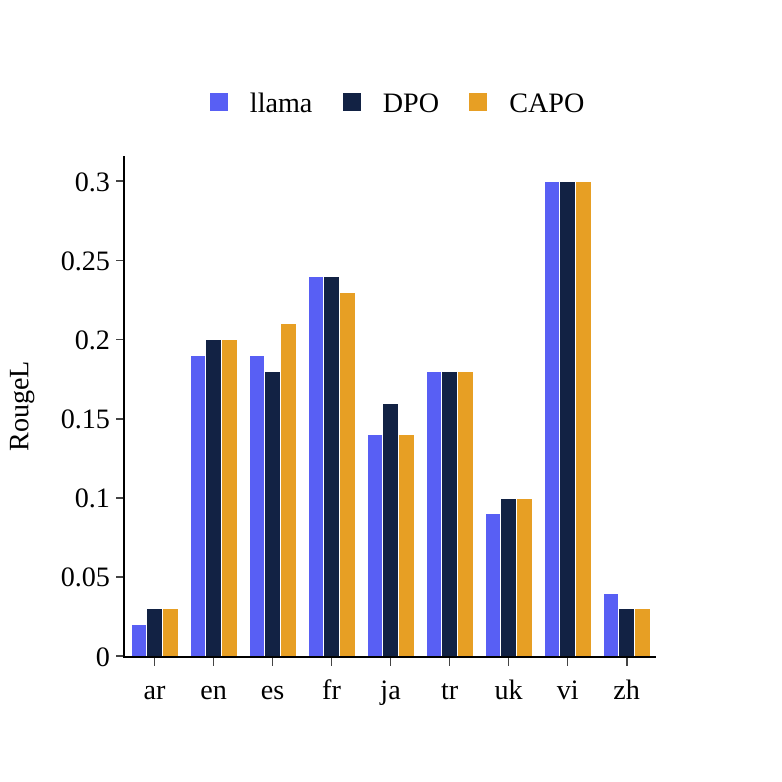}
    \caption{Average RougeL scores on XLSum dataset per language.}
    \label{fig:xlsum_rougel_perlang}
\end{figure}

\noindent \textbf{XLSum - RougeL}
Figures~\ref{fig:xlsum_rougel_perlang} present Rouge-L scores on the XLSum dataset for summarization across multiple languages. CAPO achieves comparable average performance to DPO, both slightly outperforming the base LLaMA model. The per-language breakdown in Figure~\ref{fig:xlsum_rougel_perlang} reveals that CAPO shows more consistent or improved performance in several individual languages, except for \textit{fr} and \textit{ja}. The limited improvement may be due to the limitations of ROUGE-L, particularly when summaries are not too different at surface level. As suggested in prior work~\cite{sumeval2021, deutsch-etal-2021-statistical, he-etal-2023-blind}, automatic metrics ROUGE-L or BERTScore often fails to capture human preferences related to consistency, relevance, and fluency. Because ROUGE relies on lexical overlap, it is sensitive to exact word matches and cannot effectively detect semantic equivalence or paraphrasing. That is why we additionally report win rates to better compare against human preferences as suggested in \citet{wang-etal-2024-large-language-models-fair}. 


\section{Language List}
Here, we provide the list of languages along with their corresponding language codes.

\begin{table}[ht]
\centering

\begin{tabular}{lc}
\toprule
\textbf{Code} & \textbf{Language} \\
\midrule
Arabic         & ar \\
Chinese        & zh \\
Dutch          & nl \\
English        & en \\
French         & fr \\
German         & de \\
Italian        & it \\
Japanese       & ja \\
Spanish        & es \\
Turkish        & tr \\
Ukrainian      & uk \\
Vietnamese     & vi \\
\bottomrule
\end{tabular}
\caption{Language codes and their corresponding full language names.}
\label{tab:language-codes}
\end{table}

\section{More Results}
\label{sec:app_results}

\subsection{Reward Accuracy}
\label{sec:app_reward_acc}
\begin{figure}[!t]
    \centering
    \includegraphics[trim={0 1.5cm 0 1.3cm},clip,width=0.4\textwidth]{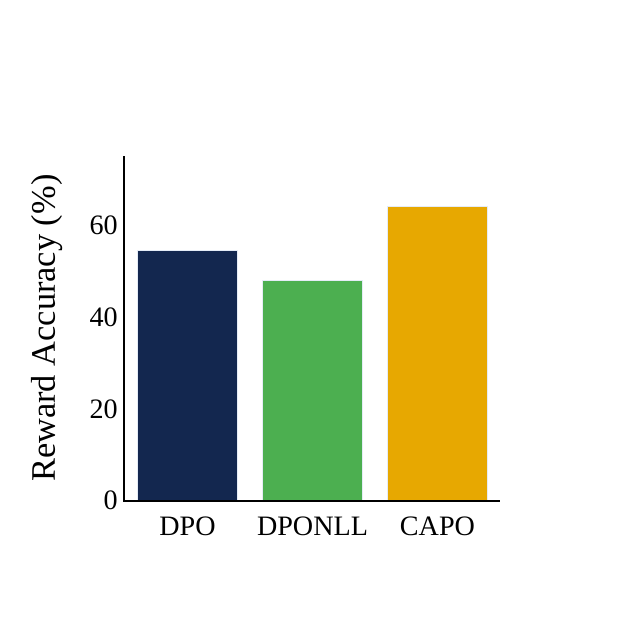}
    \caption{Comparison of average reward accuracy across different objectives on the validation data.} 
    \label{fig:reward_acc}
\end{figure}

Figure~\ref{fig:reward_acc} presents the overall average reward accuracy results.  CAPO outperforms DPO and DPONLL by 16\% and 33\% respectively.

\subsection{XLSum}
\label{sec:app_xlsum}
Tables~\ref{tab:xlsum_sd_capo_vs_llama} and~\ref{tab:xlsum_sd_capo_vs_dpo} report the standard deviation between the two runs of XLSum. Except for es, the standard deviation remains low.

\begin{table}[!t]
\centering
\begin{tabular}{crrrr}
\toprule
\textbf{Lang.} &\textbf{SD (Win)} &\textbf{SD (Tie)} &\textbf{SD (Lose)} \\
\midrule
en &1.41 &0 &1.41 \\
zh &5.66 &2.83 &2.83 \\
vi &2.83 &0 &2.83 \\
ar &1.41 &0 &1.41 \\
uk &1.41 &1.41 &0 \\
tr &0 &0 &0 \\
fr &1.41 &1.41 &2.83 \\
ja &2.83 &0 &2.83 \\
es &4.24 &0 &4.24 \\
\bottomrule
\end{tabular}
\caption{Standard deviation between two runs of XLSum win rates for CAPO vs llama.}
\label{tab:xlsum_sd_capo_vs_llama}
\end{table}

\begin{table}[!t]
\centering
\begin{tabular}{crrrr}\toprule
\textbf{Lang.} &\textbf{SD (Win)} &\textbf{SD (Tie)} &\textbf{SD (Lose)} \\\midrule
en &1.41 &0 &1.41 \\
zh &4.24 &0 &4.24 \\
vi &4.24 &0 &4.24 \\
ar &5.66 &0 &5.66 \\
uk &1.41 &0 &1.41 \\
tr &1.41 &1.41 &2.83 \\
fr &1.41 &0 &1.41 \\
ja &1.41 &0 &1.41 \\
es &1.41 &1.41 &15.56 \\
\bottomrule
\end{tabular}
\caption{Standard deviation between two runs of XLSum win rates for CAPO vs DPO.}
\label{tab:xlsum_sd_capo_vs_dpo}
\end{table}